\documentclass[10pt,twocolumn]{article} 
\usepackage{simpleConference}
\usepackage{times}
\usepackage{graphicx}
\usepackage{amssymb}
\usepackage{tikz}
\usepackage{times}
\usepackage{epsfig}
\usepackage{graphicx}
\usepackage{amsmath}
\usepackage{algorithm}
\usepackage{algorithmic}
\usepackage{float}
\usepackage{bm}
\usepackage{pgf}
\usetikzlibrary{intersections,chains,scopes,fit,trees,quotes,calc,fadings,positioning,calc,patterns}
\usepackage{hyperref}

\newcommand{\eqn}[2]
{
	\begin{equation}#1
	#2
	\end{equation}
}

\newcommand{\etbl}[4]
{
	\begin{table*}[htbp]
		\centering
		\caption{#2}
		#1
		\begin{tabular}{#3}
			#4
		\end{tabular}
	\end{table*}
}

\newcommand{\fgr}[4]
{
	\begin{figure}[!htp]
		\centering
		\includegraphics[angle=0, width=#3\textwidth]{#4}
		\caption{#2}#1
	\end{figure}
}

\newcommand{\sn}[1]
{
	\section{#1}
}

\newcommand{\ssn}[1]
{
	\subsection{#1}
}

\newcommand{\tf}[1]
{
	\textbf{#1}
}

\newcommand{\tm}[1]
{
	\textrm{#1}
}

\begin{document}

\title{Beyond Attributes: Adversarial Erasing Embedding Network for Zero-shot Learning}
\author{Xiao-Bo Jin \\
	Henan University of Technology\\
	xbjin9801@gmail.com 
	  \and Kai-Zhu Huang \\ Xian Jiaotong-Liverpool University \and Jianyu Miao  \\Henan University of Technology
}

\maketitle
\thispagestyle{empty}

\begin{abstract}
	In this paper, an adversarial erasing embedding network with the guidance of high-order attributes~(AEEN-HOA) is proposed for going further to solve the challenging ZSL/GZSL task. AEEN-HOA consists of two branches, i.e., the upper stream is capable of erasing some initially discovered regions, then the high-order attribute supervision is incorporated to characterize the relationship between the class attributes. Meanwhile, the bottom stream is trained by taking the current background regions to train the same attribute. As far as we know, it is the first time of introducing the erasing operations into the ZSL task. In addition, we first propose a class attribute activation map for the visualization of ZSL output, which shows the relationship between class attribute feature and attention map. Experiments on four standard benchmark datasets demonstrate the superiority of AEEN-HOA framework.
\end{abstract}

\section{Introduction}
Zero-shot learning (ZSL) task, first proposed in \cite{palatucci_zero-shot_2009,lampert_learning_2009} as a popular problem, currently,  regains the prevalent attention 
\cite{akata_label-embedding_2013,xian_zero-shot_2017,akata_evaluation_2015}. Unlike supervised classification task, where the label set of test images is the same as that of training images,  the label sets of training and test images are disjoint with each other in ZSL, e.g., given the images from zebra and tiger for training, and the test images are from giraffe. To make ZSL possible, the description w.r.t. training/test classes should be collected, from which it is desirable that some common informations (concepts), such as attributes \cite{farhadi_describing_2009} are extracted and served as the bridge for connecting training and test classes. Other widely used descriptions include word2vector  \cite{socher_zero-shot_2013} and sentences \cite{reed_learning_2016}. Among these descriptions, attribute is the most widely used one, in this paper, we leverage the attribute descriptions for evaluation.            
\begin{figure}[t]
	\centering
	\includegraphics[width=8.4cm]{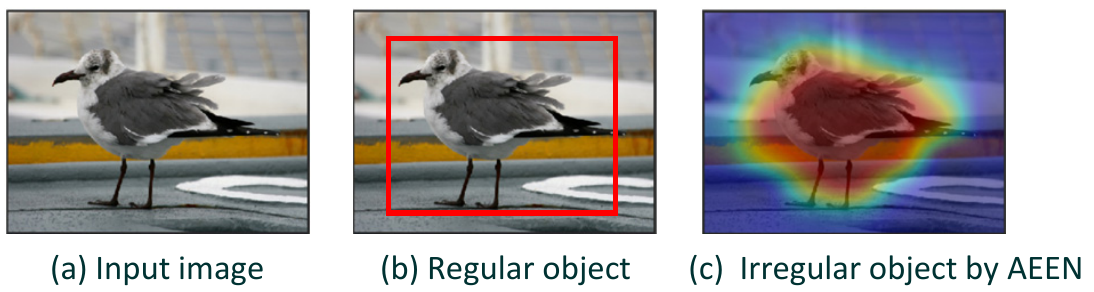}
	\caption{(a) Input image. (b) Regular object discovered by previous methods. (c) Irregular object discovered by our adversarial erasing embedding network.}
	\label{Fig1}
\end{figure}

By further projecting these descriptions onto the semantic space, we can obtain the semantic vector of each class, and then semantic vectors serve as the prototype for afterward classification on test images. A typical scenario for ZSL is thus focusing on establishing the correlation between the training/test class images and the counterpart semantic vectors; Specifically, a mapping should be learned by feeding the training class image (semantic vector) as inputs, and the output is the mapping weights, based on which, the matching scores of unseen test images with the unseen class semantic vectors can be achieved. To learn such image-semantic mapping (embedding), existing works usually design a complex optimization objective, equipped with various regularizations. This series of representative methods are based on matrix optimization  \cite{kodirov_unsupervised_2015,ye_zero-shot_2017,long_zero-shot_2018,zhang_zero-shot_2015,zhang_zero-shot_2016,romera-paredes_embarrassingly_2015,kodirov_semantic_2017,qiao_less_2016}. Moreover, triggered from the success of convolutional neural network (CNN) models \cite{he_deep_2016} on the ImageNet \cite{krizhevsky_imagenet_2012} classification task, some recent approaches resort to the CNN model to find solutions for ZSL. Li et al. \cite{li_discriminative_2018} propose to adopt zoom net \cite{fu_look_2017} for discovering global object bounding box, other CNN based methods \cite{morgado_semantically_2017,li_discriminative_2018,chen_zero-shot_2017,frome_devise:_2013,zhang_learning_2017} also take the global images as input. In addition, some specific network regularizations, such as semantically consistent regularization \cite{morgado_semantically_2017}, are incorporated into the CNN training phase. 

Few of the above approaches have considered the irregular image discovery for ZSL. Deep CNN feature based methods usually feed the whole image as inputs for extracting features. As shown in Figure \ref{Fig1}, some CNN based approaches lean to focus on the regular object box (Figure \ref{Fig1}(b)), while our proposed approach aims to discover the irregular object region (Figure \ref{Fig1}(c)). In this way, the object itself can be discovered and the background irregular regions are surppressed. The irregular object region, in some sense, corresponds to the attributes, which in turn guide the discovery of the irregular region. There are two drawbacks of human-defined attributes: (1) they are usually coarsely defined, leading that the same attribute usually corresponds to different image regions, e.g., the legs of tiger and zebra are appearently different, whilst, the same attribute ``leg'' of them can not reflect such difference; (2) multiple attributes from different classes are usually shared, resulting in the semantic vectors less discriminative, e.g., tiger and bobcat have too many identical attributes, therefore, they are hard to be distinguished. Recently, learning latent attributes \cite{li_discriminative_2018} has progressively attracted attention from ZSL community. However, they are all first-order projection based approaches, i.e., the learned attributes are the non-linear/linear combination of original ones.        

In this paper, we propose to mitigate the above problems in ZSL, by irregular image discovery and high-order attribute construction. Specifically, as shown in Figure \ref{fig:framework}, we propose an end-to-end adversarial erasing embedding network with high-order attributes (AEEN-HOA) which is designed based on user-defined attribute. AEEN-HOA targets at discovering more elaborate, diversity and discriminative high-order semantic vector for each class. The construction of high-order semantic vector is simple yet effective. To be more specific, given an input semantic vector $\bm{x}\in R^{C\times 1}$~(quantized from attributes), we first calculate the high-order correlation matrix as $M = \bm{x}\times \bm{x}^{T}\in R^{C\times C}$, then Gaussian random projection is leveraged to project $M$ onto the high-order attribute space (Figure \ref{fig:high-order-attr}).

We summarize our contributions as follows:

1) The adversarial erasing mechanism to automatically discover irregular object regions for ZSL is proposed. It is the first attempt of adopting adversarial erasing to ZSL/GZSL. 

2) To capture high-order attribute information, Gaussian random projecting (GRP) is proposed to construct the high-order attribute, which in turn can guide the irregular region discovery. To the best of our knowledge, incorporating high-order attribute into ZSL/GZSL is the first time.

3) We first propose a class attribute activation map for the visualization of ZSL output, which shows the relationship between class attribute features and attention map, which helps us understand how ZSL works.

\section{Related Works}
\textbf{Zero-shot Learning.} Direct attribute prediction (DAP) model, a seminal work for ZSL, is proposed by Lampert et al. \cite{lampert_learning_2009}. In DAP, the probabilistic attribute classifiers are first learned for each attribute, then the posteriors of the test classes are calculated for a given image. The final class is obtained by maximizing the posterior estimation. Meanwhile, multi-class classifier on seen classes for indirect attribute prediction (IAP) \cite{lampert_learning_2009} is trained. According to the scores of these seen classes, the attribute posteriors are deduced. Both DAP and IAP ignore the correlations between different attributes, a random forest approach is further introduced by \cite{jayaraman_zero-shot_2014}. 

Recently, to further construct the relationships between image and semantic vector, embedding based methods are emerging and gradually leading the ZSL community. Typically, to learn the bilinear compatibility matrix,  ALE \cite{akata_label-embedding_2013} and DEVICE \cite{frome_devise:_2013} optimize a hinge ranking loss, and SJE \cite{akata_evaluation_2015} proposes to optimize structured SVM loss. Moreover, ESZSL \cite{romera-paredes_embarrassingly_2015} and  SAE \cite{kodirov_semantic_2017} utilize the least square loss to learn the embedding matrices, and some specially designed regularizations are also incorporated. LATEM \cite{xian_latent_2016} is further proposed for extending the linear embedding methods to non-linear bilinear formulation. Other non-linear embedding methods include
CMT \cite{socher_zero-shot_2013} which is a two-layer neural network model for mapping image feature space to the semantic space, and DEM \cite{zhang_learning_2017} which projects semantic vectors of classes into the visual feature space. Besides direct projection between images and their semantic vectors, both of which are projected into some intermediate space is another group of methods for ZSL, e.g., JLSE \cite{zhang_zero-shot_2016} and SSE \cite{zhang_zero-shot_2015}, a more thorough review on ZSL is in \cite{xian_zero-shot_2017}. The above methods mainly utilize deep features, which are based on end-to-end deep CNN. The representative works are LDF \cite{li_discriminative_2018} which learn to focus regular objects, and RN \cite{yang_learning_2018} that learns to discover the relation between different images.

As for latent attribute learning, there merely exist several linear transformation methods including JSLA \cite{peng_joint_2016}, LDF \cite{li_discriminative_2018} and LAD \cite{jiang_learning_2017}, all of which are obtained by directly/indirectly regulating the inter-class and intra-class distances, and they are first-order attribute methods.

\textbf{Generalized ZSL.}   
If images from both seen and unseen classes are considered during the testing phase, ZSL becomes generalized ZSL (GZSL), which is first proposed by \cite{scheirer_toward_2013}. Then, new split for the training and test data for GZSL is proposed by \cite{xian_zero-shot_2017}. Following the new split, samples from both seen and unseen classes are utilized to conduct GZSL evaluation. 

\textbf{Adersarial Erasing Learning.} 
Adversarial Erasing aims at discovering irregular object locations, which is first proposed in \cite{wei_object_2017} for semantic segmentation task, and has been successfully applied to related fields such as object detection \cite{zhang_adversarial_2018}. Motivated by the ability of adversarial erasing learning for discovering irregular objects, We adopt the adversarial erasing to leverage ZSL, which is the first trail of using erasing learning for ZSL.

\section{Proposed Approach}

\begin{figure*}[!htp]
	\label{fig:framework}
	\centering
	\begin{tikzpicture}
	
	\tikzset{
		pics/conv/.style args={#1/#2}{
			code={
				\pgfmathsetmacro{\cubex}{#1}
				\pgfmathsetmacro{\cubey}{#2}
				\pgfmathsetmacro{\cubez}{#1}
				\draw[fill=blue!20,yshift=\cubey*0.485 cm] (0,0,0) -- ++(-\cubex,0,0) -- ++(0,-\cubey,0) -- ++(\cubex,0,0) -- cycle;
				\draw[fill=blue!20,yshift=\cubey*0.485 cm] (0,0,0) -- ++(0,0,-\cubez) -- ++(0,-\cubey,0) -- ++(0,0,\cubez) -- cycle;
				\draw[fill=blue!20,yshift=\cubey*0.485 cm] (0,0,0) -- ++(-\cubex,0,0) -- ++(0,0,-\cubez) -- ++(\cubex,0,0) -- cycle;
			}
		}
	};
	
	\tikzset{
		pics/map/.style args={#1/#2}{
			code={
				\fill[blue!#2,yshift=-1.1cm](0,0) rectangle +(#1,#1);
				\fill[red!#2,yshift=-1.1cm](0.1,0.1) rectangle +(#1,#1);
				\fill[green!#2,yshift=-1.1cm](0.2,0.2) rectangle +(#1,#1);
				\fill[yellow!#2,yshift=-1.1cm](0.3,0.3) rectangle +(#1,#1);
			}
		}
	};

	\tikzset{relu/.pic={
			\draw[blue,thick,yshift=-0.8cm] (0,0) -- (2cm,0);
			\draw[blue,thick,yshift=-0.8cm] (0,0) -- (135:2cm);
		}
	};
	
	\tikzset{threshold/.pic={
			\draw[blue,thick,xshift=-1.5cm] (0,0) -| (1,1) -| (2,0) -- (3,0);
		}
	};
	
	\tikzset{semantic/.pic={
			\node[fill=blue,path fading = west,rectangle,minimum height = 0.15cm,minimum width = 1.5cm](class1){};
			
			\node[fill=blue,path fading = west,rectangle,minimum height = 0.15cm,minimum width = 1.5cm,above = 0.1cm of class1.north] (class2){};
			
			\node[left = 0.1cm of class1.west]{\footnotesize project};

			\node[fill=red,path fading = west,rectangle,minimum height = 0.15cm,minimum width = 1.5cm,below = 0.1cm of class1.south] (class3){};
		}
	};
	
	\tikzset{pred/.pic={
			\filldraw[draw = blue,fill=blue!40,yshift = -1.12cm] plot[xbar,bar width = 0.6cm]
			coordinates{(1,0) (0.4,0.75) (1.7,1.5) (1.6,2.25)};
		}
	};

	\begin{scope}[start chain = 1 going above,node distance=0.3cm]
	\node[draw,inner sep=0,on chain = 1] (A)  {\includegraphics[width=1cm]{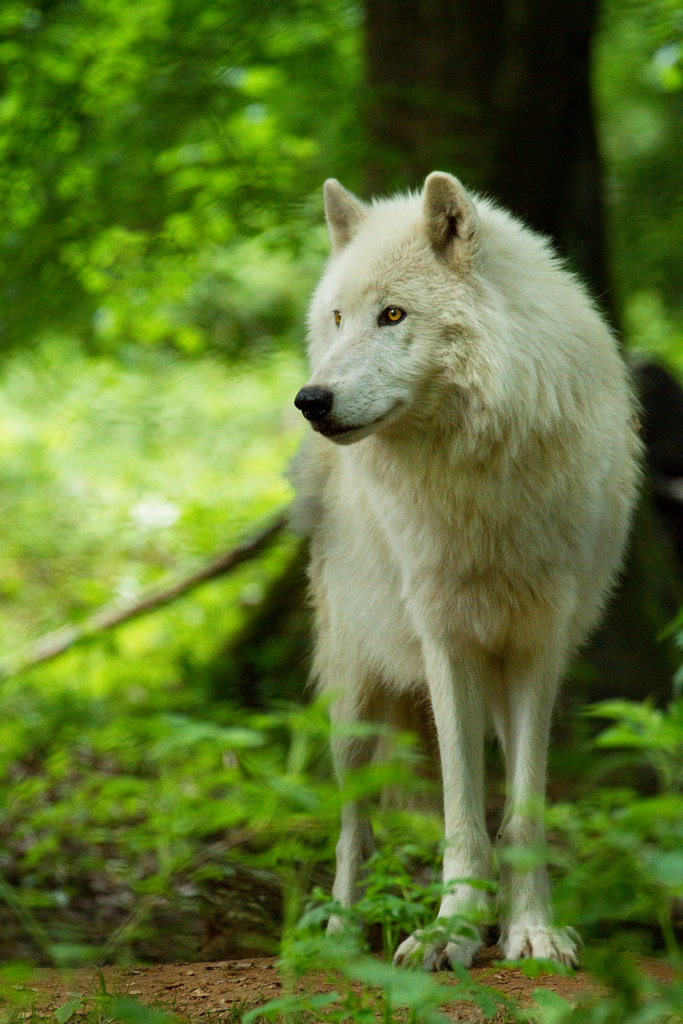}};
	\node[draw,inner sep=0,on chain = 1] (B)  {\includegraphics[width=1cm]{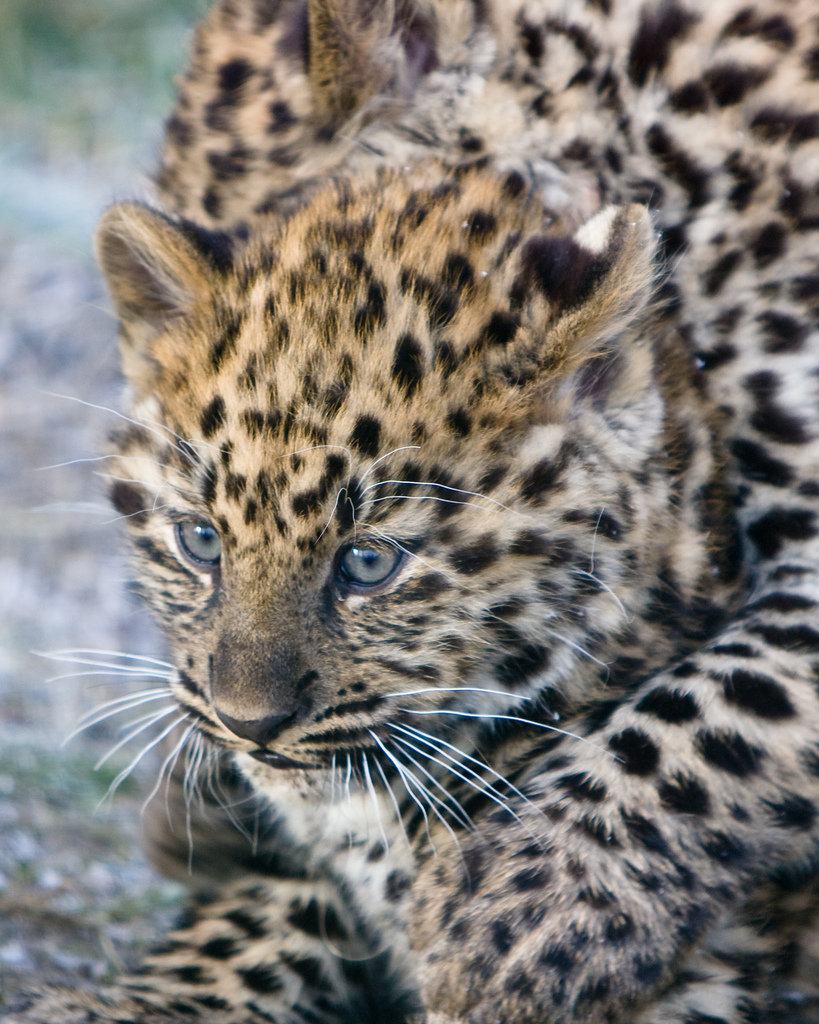}};
	\node[draw,inner sep=0,on chain = 1] (C) {\includegraphics[width=1cm]{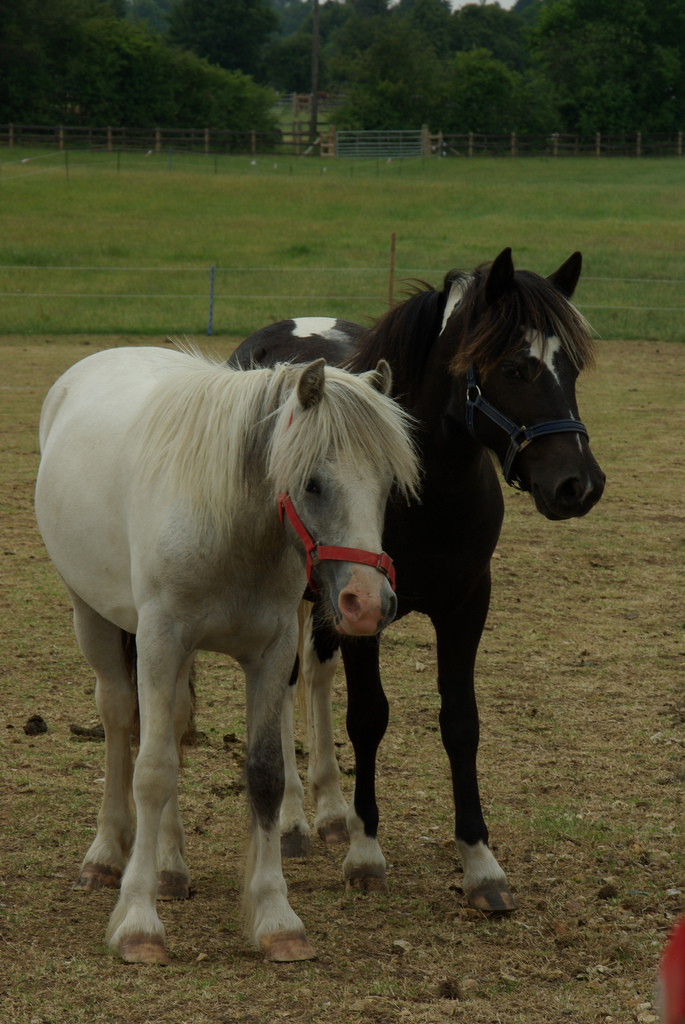}};
	\end{scope}
	
	\node[draw=blue,ultra thick,dotted,fit=(A)(B)(C)](imgs){};
	\node[below=0.2cm of imgs.south]{\footnotesize input images};
	
	\begin{scope}[start chain = 2 going right,
	node distance=0.2cm,every node/.style={minimum width=0.25cm,fill=yellow!40}]
	\node[on chain = 2,right=0.7cm of imgs.east,fill,rectangle,minimum height=3cm](C){};
	\node[on chain = 2,fill,rectangle,minimum height=2.5cm](D){};
	\node[on chain = 2,fill,rectangle,minimum height=2cm](E){};
	\end{scope}
	
	\node[draw=blue,ultra thick,dotted,fit=(C)(D)(E)](cnnmodels){};
	\node[below=0.2 cm of cnnmodels.south]{\footnotesize resnet101};
	
	\draw[-latex,ultra thick](imgs) -- (cnnmodels);
	
	\draw pic[scale=0.5,right=0.75cm of cnnmodels.east,local bounding box=conv]{conv=0.5/8};
	
	\node[below=0.2cm of conv.south]{\footnotesize conv$_{3 \times 3}$};
	
	\draw pic[scale=0.8,right=0.7cm of conv.center,local bounding box=output]{map=2/50};
	\node[below = 0.2cm of output]{\footnotesize conv output};
	
	\draw[-latex,ultra thick](cnnmodels.east) -- (conv);
	\draw[-latex,ultra thick](conv.east) -- (output);
	
	\node at (output.east){}
	[edge from parent fork right,grow=right,sibling distance = 3.5cm,level distance = 1cm,line width = 1.8pt]
	child {node (branch1) {}}
	child {node (branch2) {}};
	
	\draw pic[scale=0.3,right=0.01cm of branch2,local bounding box=relu]{relu};
	
	
	\node[below=0.2cm of branch2.south east]{\footnotesize relu};
	
	\draw pic[scale=0.35,above=0.6cm of output.north,local bounding box=threshold]{threshold};
	\node[above right = 0.01cm of threshold.north]{\footnotesize threshold};

	\draw[-latex,very thick](output.north) -- (threshold.south);
	
	\node[above=1.5cm of relu.north west](mask){};
	\draw[-latex,very thick] (threshold.north)+(0,0.1cm) |- (mask);
	\draw pic[scale=0.8,right=0cm of mask.center,local bounding box=maskoutput]{map=2/30};

	\node[draw=red!80,ultra thick,dotted,fit=(threshold)(maskoutput)(relu)](erase){};
	\node[left=0.2cm of erase,red!80]{C-ReLU};
	
	\node[right=0.8cm of relu.east](joint){};
	\draw[very thick,-latex] (maskoutput.east) -| (joint.center) node[label={above right:$\times$}]{};
	
	\begin{scope}[start chain = 3 going right,node distance=0.3cm]
	\node[right=-0.4cm of relu.east,on chain = 3](H){};
	\node[right=0.01cm of H,on chain = 3](J){};
	\end{scope}
	
	\draw pic[scale = 0.5,right=1.6cm of J.center,local bounding box = conv1]{conv=0.2/4};
	\draw[ultra thick,-latex] (relu.east) -- (conv1);
	\node (conv2) at (branch1 -| conv1) {};
	
	\draw pic[scale=0.5,right=0cm of conv2.center,local bounding box=cov2]{conv=0.2/4};
	\draw[ultra thick,-latex] (branch1.west) -- (conv2);
	\node[below = 0.2cm of cov2]{\footnotesize conv$_{1 \times 1}$};
	
	\node[right = 0.5cm of conv1,rectangle,fill=green!40,minimum height = 1cm,minimum width = 0.8cm](max1){};
	
	\node[right = 0.5cm of conv2,rectangle,fill=green!40,minimum height = 1cm,minimum width = 0.8cm](max2){};
	\node[below= 0.2cm of max2]{\footnotesize max$_{5 \times 5}$};
	
	\draw[-latex,ultra thick] (cov2) -- (max2);
	\draw[-latex,ultra thick] (conv1) -- (max1);
	
	\node (sem) at ($(max1)!0.5!(max2)$){};
	\draw pic[scale=0.5,right=0.5cm of sem.east,local bounding box=semantic]{semantic};
	
	\draw[-latex,line width=2pt] (max1) -| (semantic);
	\draw[-latex,line width=2pt] (max2) -| (semantic) node [pos=0.8,label={right:\footnotesize high-order attr}]{};
	
	\node (output)[right = 0.9cm of semantic.east]{$\oplus$};
	\node [left = 0.03cm of output.center] {\footnotesize avg};
	
	\draw[-latex,ultra thick] (max1) -| (output);
	\draw[-latex,ultra thick] (max2) -| (output);
	
	\draw pic[scale=0.5,right=0.8cm of output.east,local bounding box=pred]{pred};
	\node[below=0.2cm of pred]{\footnotesize predict};
	
	\draw [-latex,ultra thick] (output) -- (pred);
	
	\end{tikzpicture}
	\caption{Overview of the proposed AEEN-HOA approach. AEEN consists of two branches after a shared backbone network (e.g. resnet101). The structure of the lower branch is a convolutional network with the size $1 \times 1$ followed by a maximum pooling layer, and the upper branch is similar to that of the bottom one, except a C-ReLU layer which is inserted in front. The outputs of both branches are projected onto the space spanned by the high-order attributes.}
\end{figure*}
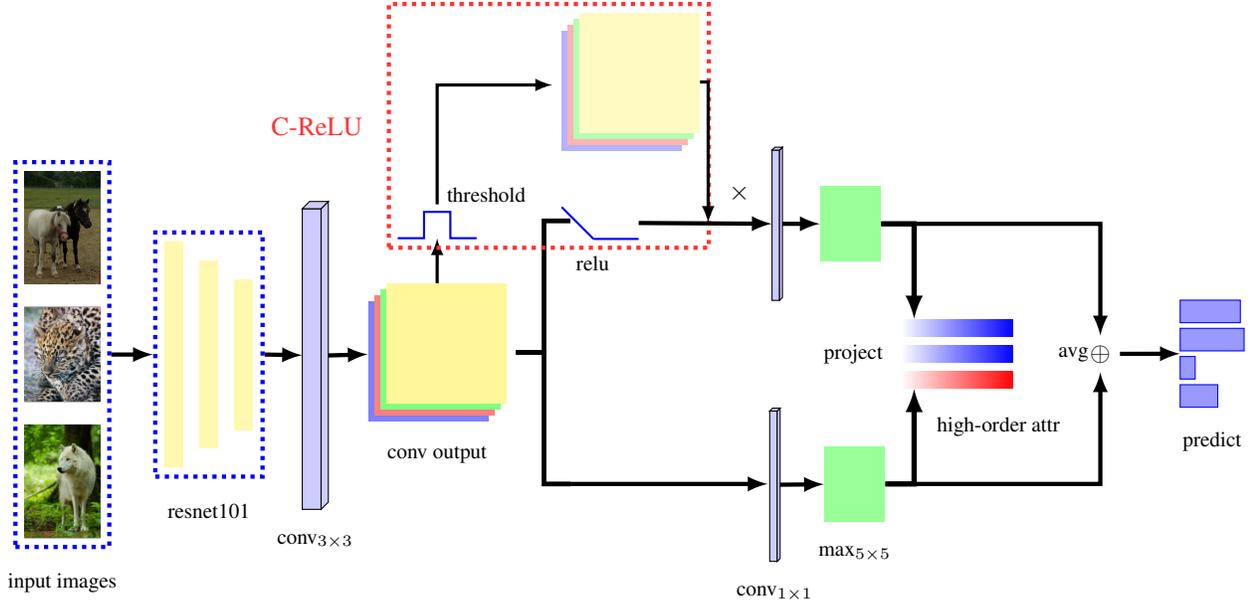

We are given a set of source classes $C_{S} = \{l_1,l_2,\cdots,l_s\}$ and $N$ labeled source samples $D = \{(\bm{I}_i,y_i)\}_{i = 1}^N$ for training, where $\bm{I}_i$ is the $i$-th training image and $y_i(y_i \in C_S)$ is its label. Given a new test image $\bm{I}_j$, the goal of ZSL is to assign it to an unseen class label which is from $C_{U} = \{l_{s + 1},\cdots,l_{s + u}\}$. Note that the label sets from the training (seen) classes and the test (unseen) classes are disjoint from each other, i.e., $C_{S} \cap C_{U} = \phi$. Each class label $y$ (both seen/unseen classes) is associated with a predefined semantic vector $\varphi(y)$. 

\subsection{Adversarial Erasing Embedding Network}
Adversarial Erasing  Network (AEN) \cite{hou_self-erasing_2018} is an extension of class activation maps (CAM) \cite{zhou_learning_2015}, where fully connected layers can aggregate the features of the last convolutional layer for the localization purpose. Therefore, AEN can essentially discover the irregular objects. These irregular objects can assist the ZSL tasks and so we propose to embed AEN for ZSL tasks, which is an end-to-end adversarial erasing embedding network framework (AEEN).

AEEN consists of two branches after a shared backbone network (e.g. resnet101). The structure of the lower branch is a convolutional network with the size $1 \times 1$ followed by a maximum pooling layer, and the upper branch is similar to that of the bottom one, except a C-ReLU layer which is inserted in front.

We consider a fully convolutional network (FCN) and denote the last convolutional feature maps by $S_{K \times H \times H}$, where $H \times H$ is the spatial size and $K$ is the number of channels. We aggregate the feature map $S$ with $C$ groups of weights to obtain $C$ weighted feature maps called as the localization map $L_{c}, c = 0,1,\cdots, C - 1$, which can be computed as 
\eqn{}{
	L_{c} = \sum_{k = 0}^{K - 1} S_{k} \cdot W_{k,c},
}
where $S_i$ is the $i$-th channel of feature map with the size $H \times H$. The above localization can be implemented by a convolutional layer with the kernel size $1 \times 1$ (see $\tm{conv}_{1 \times 1}$ unit of Figure \ref{fig:framework}). 

As shown in the red block diagram of Figure \ref{fig:framework}, we introduce the erase operation to learn to highlight the attention map, where C-ReLU function merges a binary mask with the ReLU function.  C-ReLU is defined as
\eqn{}{
	\tm{C-ReLU}(x) = \max(x,0)\cdot \theta_{\delta}(x),
} 
where $\theta_{\delta}(x)$ is a binary mask: $\theta_{\delta}(x) = 1$ if $x \ge \delta$, and $\theta_{\delta}(x) = -1$ otherwise. In our work, we set a parameter $\delta_{k}$ for each channel $S_{k}$ ($k = 0,1,\cdots,K - 1$) of the feature map. 

\subsection{Extraction of High-order Features}
Most previous works on learning latent attributes in ZSL focus on the class attribute itself or its linear/non-linear transformation, such as the form of two-layer neural network
\eqn{}{
	f(\varphi(y)) = f_2(A_2 f_1(A_1 \varphi(y) + b_1) + b_2),
}
where $\varphi$ is a predefined semantic vector of the class $y$.

\begin{figure}[H]
	
	\centering
	\begin{tikzpicture}[scale=0.6,transform shape]
	\node(A)[rectangle, draw,inner sep=0,xshift=-0.5cm]
	{
		\tikz{\draw[step=5mm] (0,0)  grid (0.5,3);
			\fill (0,0.5*2) rectangle +(0.5,0.5);
			\fill (0,0.5*5) rectangle +(0.5,0.5);
		}
	};
	\node[draw=none,above=0.3cm of A]{$n \times 1$};
	
	\node(B)[right=0.1cm of A]{$\times$};
	\node(C) [rectangle, draw,inner sep=0,transform shape,rotate=90,right=1.5cm of B.south east]
	{
		\tikz{\draw[step=5mm] (0,0)  grid (0.5,3);
			\fill (0,0.5*1) rectangle +(0.5,0.5);
			\fill (0,0.5*4) rectangle +(0.5,0.5);
		}
	};
	
	\node[draw=none,above right=0.3cm of C]{$1 \times n$};
	
	\draw (C)+(1.5cm,0) node[anchor=west]{$=$};
	
	\node (D)[rectangle, draw,inner sep=0,right=2.3cm of C.east,yshift=-0.25cm]
	{
		\tikz{\draw[step=5mm] (0,0)  grid (3,3);
			\fill (0.5*1,0.5*5) rectangle +(0.5,0.5);
			\fill (0.5*4,0.5*5) rectangle +(0.5,0.5);
			
			\fill (0.5*4,0.5*2) rectangle +(0.5,0.5);
			\fill (0.5*1,0.5*2) rectangle +(0.5,0.5);
		}
	};
	
	\node[draw=none,above=0.3cm of D]{$n \times n$};
	
	\draw[-latex,line width=1pt](D.east)+(0.2cm,0)-- +(1cm,0) node (E)[above,midway]{vec};
	
	\draw (E)+(0.8cm,2cm) node(F)[rectangle, draw,inner sep=0]
	{
		\tikz{\draw[step=5mm] (0,0)  grid (0.5,1.5);
			\fill (0,0.5*1) rectangle +(0.5cm,0.5cm);
		}
	};
	
	\node[draw=none,above=0.3cm of F]{$n^2 \times 1$};
	
	\node(G)[below=of F]{$\vdots$};
	
	\node(H)[below=of G]{
		\tikz{\draw[step=5mm] (0,0)  grid (0.5,1.5);
		}
	};
	
	\draw[-latex,line width=1pt](G)++(0.5cm,-0.1cm)-- +(1cm,0) node (H)[above,midway]{GRP};
	
	\node(I)[right=0.1cm of H,yshift=-0.3cm]{
		\tikz{\draw[step=5mm,fill=blue!30] (0,0)  grid (0.5,2) rectangle (0,0);
		}
	};
	
	\node[draw=none,above=0.3cm of I]{$m$};
	
	\node(J)[right=0.9cm of I]{
		\tikz{\draw[step=5mm] (0,0)  grid (0.5,5);
			\draw[step=5mm,fill=blue!30] (0,3)  grid (0.5,5) rectangle (0,3);
		}
	};
	\draw[-latex,line width=1pt] (I) -- (J) node (R)[above,midway]{cat};
	\node[draw=none,above=0.3cm of J]{$m + n$};
	
	\end{tikzpicture}
	\caption{Merge of the high-order and original class attributes: the outer product of the class attribute is vectorized by row and then project into a reduced space to obtain a compact high-order representation.}
	\label{fig:high-order-attr}
\end{figure}
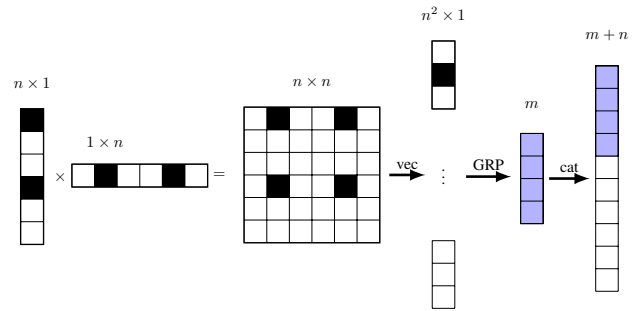

However, in many vision tasks, the relationship between the class attributes carries the relevant information, which is helpful for ZSL. We use the outer product to encode the relations between the class attributes as
\eqn{}{
	L_{y} = \tm{vec}\left(\phi(y)\cdot \phi(y)^T \right).
}
Each element in the matrix $\phi(y) \cdot \phi(y)^T$ will constitute evidence for exactly one type of shift and detect the coincidences acting like AND-gates (Figure \ref{fig:high-order-attr}). 

For the sake of faster processing time and smaller model sizes, we need an efficient way to remove the unimportant attribute relations.  Random projections show appealing properties of preserving the distance quite well. The projected onto a random lower-dimensional subspace yields comparable results to PCA yet with computationally less expensive costs \cite{bingham_random_2001}. The original d-dimensional data using a random $r \times d$ matrix $W^{RP}$ whose rows have unit lengths.  With the projection matrix $W^{RP}$, the input is mapped onto $r$ dimensions of subspace is the time complexity of $O(rdn)$. Gaussian random projection \cite{achlioptas_database-friendly_2003} projects the original input $X$ on the reduced subspace with the random matrix whose components are selected from the Gaussian distribution $N(0,1/r)$.

\ssn{AEEN with High-Order Attributes (AEEN-HOA) for ZSL problems}

After the class activation map in both of branches ($\tm{conv}_{1 \times 1}$ in Figure \ref{fig:framework}), we add $5 \times 5$ max pooling (the green square in Figure \ref{fig:framework}) and then project them into the new class attribute semantic space.

Our ZSL model aims to learn the relation between the visual feature space and the semantic space. Formally,
\eqn{}{
	F(\bm{I}_i;W) = \phi(\bm{I}_i)^T W \varphi(y) 
} 
where $W$ is the weight to learn in a fully connected layer and an image representation $\phi(\bm{I}_i)$ is mapped into the class attribute semantic space. It is similar to the classification score in traditional object recognition task, where the sum of the cross-entropy loss of two branches can be used:
\eqn{}{
	L = L_1 + L_2.
}

At the test stage, an unseen image $I_{u}$ can be assigned to the most matched class $y^{*} \in C_{U}$
\eqn{}{
	y^{*} = \mathop{\arg\max}_{l \in C_{U}} \phi(\bm{I}_u)^T W \varphi(l)
}

\sn{Experiments}
\ssn{Datasets and settings}
\tf{Datasets.} We select two fine-grained ones (CUB and SUN), and two coarse-grained datasets (AWA2 and aPY).


\tf{CUB}(Caltech-UCSD Birds-200-2011) is a medium scale dataset with respect to the number of classes and images. We follow the classes split of CUB with 150 training (50 validation classes) and 50 test classes. \tf{SUN} contains 14340 images coming from 717 types of scenes annotated with 102 attributes, where 645 classes (65 classes for validation) are chosen for the training and 72 classes for testing. \tf{AwA2} contains 37,322 images of the same 50 classes of animals for training (13 classes for validation) and another 10 classes for testing, which is an extension of \tf{AwA1}.
Finally, \tf{aPY} contains 32 classes with 64-dimension attribute vectors including 20 Pascal classes for training and 12 Yahoo classes for testing.

\tf{Implementation details.} We conduct the experiments under two kinds of ZSL settings, including the standard splitting (SS) and the proposed splitting (PS). In addition, we also give the results in the generalized ZSL, where the test samples may come from either the training classes or test classes. 

For aPY, we crop the images from bounding boxes due to multiple objects in each image. Our image embedding vectors correspond to 2048-dim top-layer pooling units of ResNet-101 network. We use the original ResNet-101 that is pre-trained on ImageNet with 1000 classes. Most of previous ZSL methods adopt the fixed pre-trained features, but we believe that it is inappropriate that regulating the image representation with fixed image features. In general, an end-to-end framework will lead to better performance. We initialize the final full connected linear layer with the attribute matrix and fix them during the training process.

SGD is used to optimize our model with a minibatch size of $64$. An initial learning rate is randomly taken from the real range $[0.0001,0.01]$. For our SGD algorithm, we use the cycling learning rate strategy, where the starting cycle is set to 10 epochs and then multiplied by a factor $2$ ($T_{mul} = 2$). Other training parameters such as the dropout rate, momentum and weight decay are set to 0.4, 0.9 and 0.0005, respectively. For the threshold used in the erase network, we set the threshold $\delta$ to $\xi$ times of the maximum value of each channel of the attention map inputted to C-ReLU layer, where $\xi$ is taken from the range $[0.001,0.1]$. For the extraction of high-order features, we set the reduced dimension to $\gamma$ times of the dimension of the original attributes, where $\gamma$ is a float number chosen from $\{0.3,4\}$.

\ssn{Fast hyper-parameters search}
Random search \cite{bergstra_random_2012} is able to find models that are as good as the grid search at less computation cost. For each configuration, the training of deep learning on large-scale datasets is the main computational bottleneck: it ofter require several days to obtain a reasonable results.

\fgr{\label{fig:learning-rate}}{Cycling learning rate and const learning rate}{0.45}{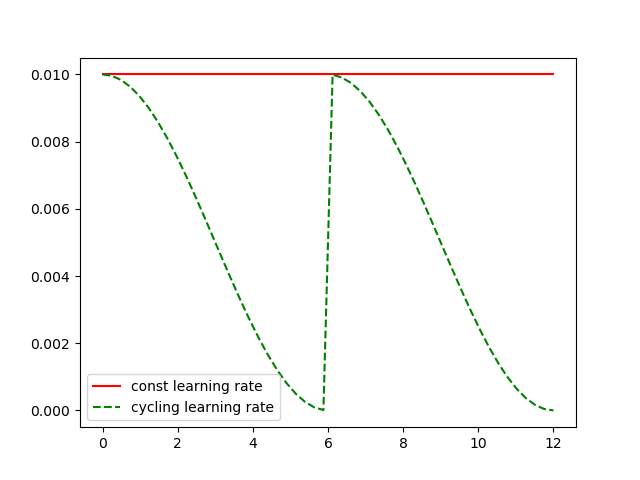}

The cycling learning rate \cite{loshchilov_sgdr:_2016} can help us achieve better performance and faster convergence rate than the constant learning rate within few epochs.
In the following, we use the cycling learning rate strategy  to search a best parameter for the ZSL problems, which simulates a new restart of SGD after $T_i$ epochs are implemented. During $T_i$ epochs, the learning value is varying from its maximum to minimum (e.g. 0). Formally, the learning rate with a cosine annealing is computed as
\eqn{}{
	\alpha = \frac{\alpha_{max}}{2}\left(1 + \cos \left(\frac{T_{cur}}{T_i}\pi \right) \right),
}
where $\alpha_{max}$ is the max learning rate, $T_{cur}$ is an accumulating epochs from the last restart. It is noted that each batch has its learning rate since $T_{cur}$ is updated during each batch iteration. Meanwhile, we increase $T_i$ by a factor of $T_{mul}$ at every restart. 

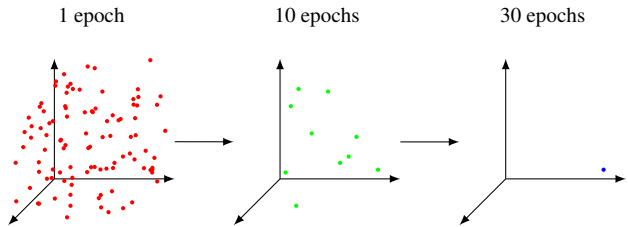
\begin{figure}[!htp]
	\label{fig:random-search}
	\centering
	\tikzset{
		pics/3dpoints/.style args={#1/#2}{
			code={
				\pgfmathsetseed{7};
				\draw[-latex] (0,0,0) -- +(0,2,0);
				\draw[-latex] (0,0,0) -- +(0,0,2);
				\draw[-latex] (0,0,0) -- +(2,0,0);
				\foreach \x in {1,...,#1}
				{
					\fill[#2] (rnd*2,rnd*2,rnd*2) circle (1pt);
				}	
			}
		}
	}
	\begin{tikzpicture}
	
	\draw pic[scale=0.8,local bounding box=large]{3dpoints=100/red};
	
	\draw[xshift=3cm] pic[scale=0.8,local bounding box=small]{3dpoints=10/green};
	
	\draw[xshift=6cm] pic[scale=0.8,local bounding box=final]{3dpoints=1/blue};
	
	\draw[-latex] (large) -- (small);
	\draw[-latex] (small) -- (final);
	
	\node[above=0.3cm of large]{\footnotesize 1 epoch};
	
	\node[above=0.3cm of small]{\footnotesize 10 epochs};
	
	\node[above=0.3cm of final]{\footnotesize 30 epochs};
	
	\end{tikzpicture}
	\caption{Fast parameter random search process consists of two phases: (1) randomly generate 100 parameter configurations, and the algorithm runs 1 epoch under each configuration; (2) select ten best configurations from 100 parameter configurations for 10 rounds; (3) finally chose a best parameter configuration among 10 candidate ones and run 30 epochs on the test set.}
\end{figure}

Given a large group of candidate parameters (e.g. 100) randomly chosen from a user-defined range, we run one epoch for each candidate parameter. According to the performance on the validation dataset, we select the top ten parameter configurations and run ten epochs to choose the best parameter configuration from these ten groups. Finally, we report the final results by running another 30 epochs on the test dataset.

In summary, our search strategy takes into account both the breadth and precision of the search. It gradually narrows the scope of the search and improves the precision of the search during the search process.

\ssn{Exploration of AEEN-HOA algorithm}
\tf{Effects of Cycling learning rate.} Figure \ref{fig:learning-rate} gives a comparison in terms of the accuracy in the first ten epochs for the constant learning rate and the cycling learning rate with different configurations. The length and multiplier of the cycle vary from $\{2,10\}$ and $\{1,1.1,1.5,2\}$, respectively. After three epochs, the constant learning rate begins to catch up with the cycling learning rate. But at the 6th epochs, the cycling learning rate overpasses the constant learning rate. In practice, the increasing period may slow down the decay speed of the learning rate. As seen from Figure \ref{fig:learning-rate},  we can obtain the best performance with the cycle multiplier $2$ and $1.5$. Our proposed algorithm achieve the highest accuracy in case of $cycle\_len = 10$ and $cycle\_mul = 2$, which verifies that it is a good empirical setting  in the deep learning \cite{loshchilov_sgdr:_2016}.

\fgr{\label{fig:learning-rate}}{Comparisons between the step learning rate and the cycling learning rate where the initial learning rate is 0.001 with the settings of different cycle lengths and the cycle multipliers}{0.5}{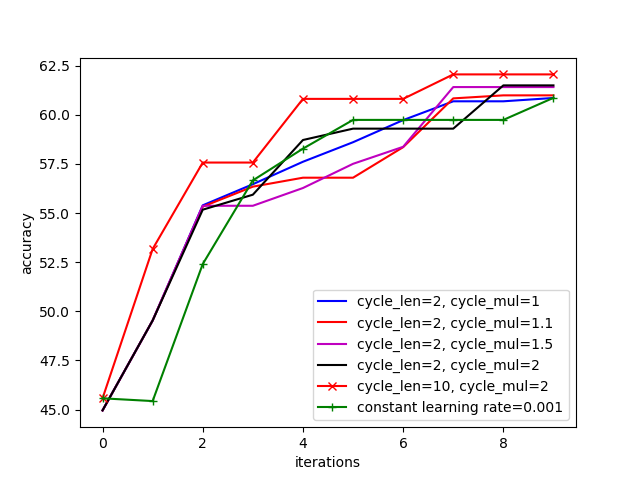}

\tf{Results on rectifying output of GZSL.}  We observe that the output in the training and test classes is not comparable. When the output from the training classes dominate, the classification performance of the training classes is higher than that of the test classes, and vice versa. We argue that the performance of the training classes is not necessarily better than one of the test classes, which can be found in Chao's work \cite{chao_empirical_2016}. 

Figure \ref{fig:train-test-gszl} demonstrates the necessity of rectifying the outputs of GZSL.
We can see that the training accuracies decrease gradually but the test accuracies increase when we put the instances one by one from the training class into the test class.  The higher the harmonic measure is, the better an algorithm is able to balance. At some point, we achieve the maximum harmonic average of the training accuracy and the test accuracy.  On the red square point (Figure \ref{fig:train-test-gszl}), the training accuracy is the most close to the test accuracy. In Table \ref{tab:gzsl}, we further validate the advantages of rectifying approaches.

\fgr{\label{fig:train-test-gszl}}{Training class-test class accuracy curve of our algorithm on SUN and CUB datasets: the red square point and the blue cross point show the training accuracy and test accuracy after and before rectifying.}{0.5}{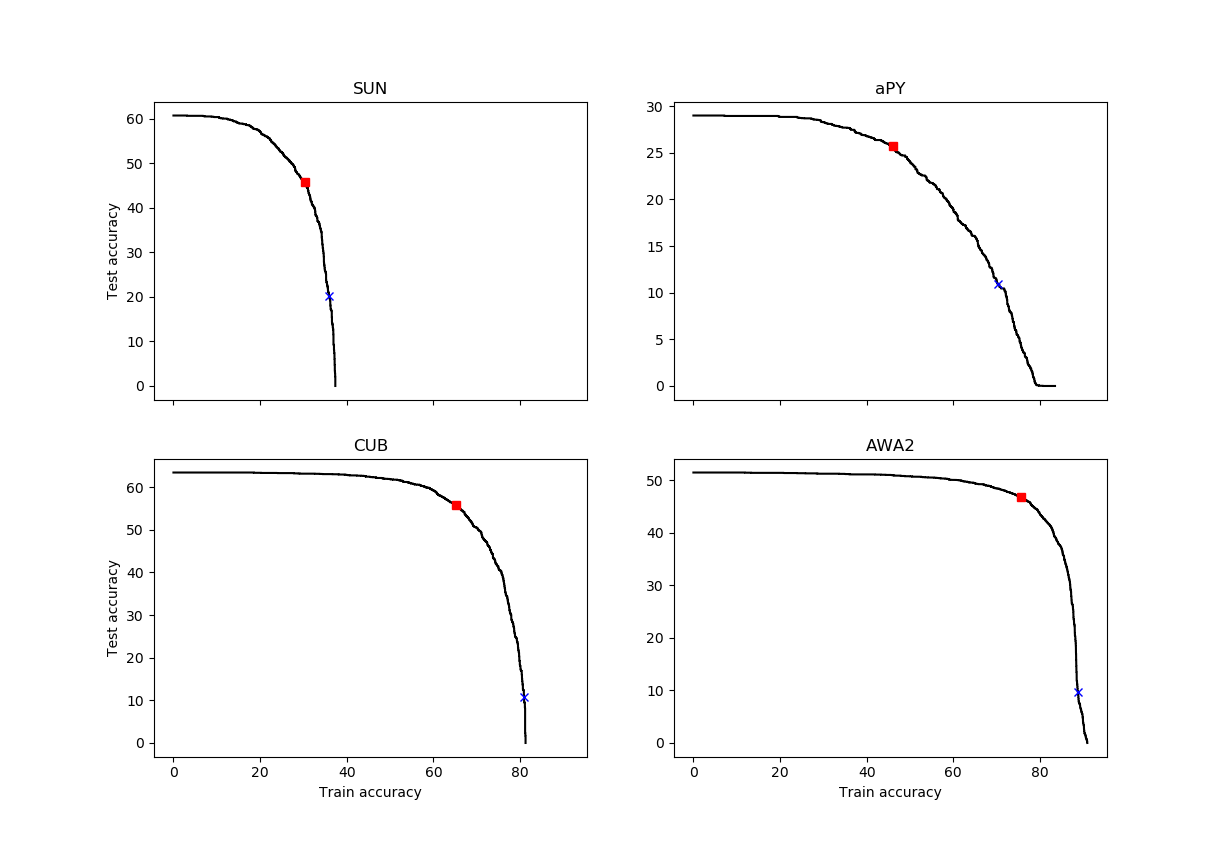}

\begin{figure}[!htp]
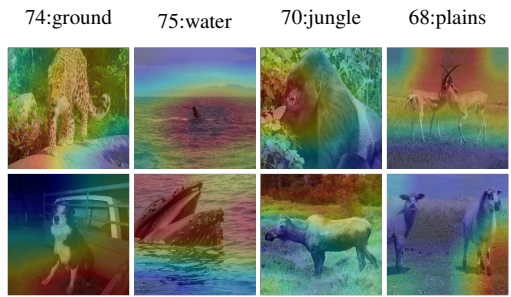

	\centering
	\begin{tikzpicture}[scale=0.8,transform shape]      
	\foreach \i/\idx/\f in {0/74/ground,1/75/water,2/70/jungle,3/68/plains}
	{
		\node[draw,inner sep=0] (0) at (\i*2.1cm,0) {\includegraphics[width=2cm]{\f-0.jpg}};
		\node[draw,inner sep=0] (1) at (\i*2.1cm,2.1cm) {\includegraphics[width=2cm]{\f-1.jpg}};
		\node[above=0.2cm of 1.north]{\idx:\f};
	}	
	\end{tikzpicture}
	\caption{Class attributes activation map of AWA2 dataset (numbering the class attributes from zero): the maps highlights the object regions related to the class attributes, e.g. ground, water, jungle and plains.}
	\label{fig:feat-heat-map}
\end{figure}

\etbl{\label{tab:results}}{Zero-shot learning results on SUN,CUB,AWA2 and aPY.}{l|cc|cc|cc|cc|cc}{
	\hline
	Method & SUN-SS & SUN-PS & CUB-SS & CUB-PS & AWA2-SS & AWA2-PS & aPY-SS & aPY-PS \\
	\hline
	DAP \cite{lampert_attribute-based_2014}&38.9&39.9&37.5&40.0&58.7&46.1&35.2&33.8\\
	IAP \cite{lampert_attribute-based_2014}&17.4&19.4&27.1&24.0&46.9&35.9&22.4&36.6\\
	CONSE \cite{norouzi_zero-shot_2013} &44.2&38.8&36.7&34.3&67.9&44.5&25.9&26.9\\
	CMT \cite{socher_zero-shot_2013} &41.9&39.9&37.3&34.6&66.3&37.9&26.9&28.0\\
	SSE \cite{zhang_zero-shot_2015} &54.5&51.5&43.7&43.9&67.5&61.0&31.1&34.0\\
	LATEM \cite{xian_latent_2016} &56.9&55.3&49.4&49.3&68.7&55.8&34.5&35.2\\
	ALE \cite{akata_label-embedding_2016} &59.1&58.1&53.2&54.9&80.3&62.5&30.9&39.7\\
	DEVISE \cite{frome_devise:_2013} &57.5&56.5&53.2&52.0&68.6&59.7&35.4&\tf{39.8}\\
	SJE \cite{akata_evaluation_2015} &57.1&53.7&55.3&53.9&69.5&61.9&32.0&32.9\\
	ESZSL \cite{romera-paredes_embarrassingly_2015} &57.3&54.5&55.1&53.9&75.6&58.6&34.4&38.3\\
	SYNC \cite{changpinyo_synthesized_2016} &59.1&56.3&54.1&55.6&71.2&46.6&39.7&23.9\\
	SAE \cite{kodirov_semantic_2017} &42.4&40.3&33.4&33.3&80.7&54.1&8.3&8.3\\
	GFZSL \cite{verma_simple_2017} &62.9&60.6&53.0&49.3&79.3&63.8&\tf{51.3}&38.4\\
	\hline
	SP-AEN \cite{chen_zero-shot_2017} & --- & 59.2 & --- & 55.4 & --- & 58.5& --- & 24.1 \\
	PSR \cite{annadani_preserving_2018} & --- & 61.4 & --- & 56 & --- & 63.8 & --- & 38.4 \\    
	\hline
	AEEN  &61.5 &60.1 &\tf{70.8} &\tf{73.5} &81.5 &64.4 &43.2 &37.2\\
	AEEN-HOA  &\tf{63.5} &\tf{62.5} &68.4 &72.2 &\tf{87.1} &\tf{67.2} &45.7 &38.3\\
	\hline
}

\etbl{\label{tab:gzsl}}{Generalized Zero-Shot Learning on Proposed Split (PS) measures including the training accuracy, test accuracy and harmonic mean.}{l|ccc|ccc|ccc|ccc}{
	\hline
	Method & \multicolumn{3}{|c|}{SUN}  & \multicolumn{3}{|c|}{CUB}
	& \multicolumn{3}{|c|}{AWA2} & \multicolumn{3}{|c}{aPY}\\
	\hline
	& tr & te & H & tr & te & H & tr & te & H & tr & te & H\\
	\hline
	DAP \cite{lampert_attribute-based_2014} &4.2&25.1&7.2&1.7&67.9&3.3&0.0&84.7&0.0&4.8&78.3&9.0\\
	IAP \cite{lampert_attribute-based_2014} &1.0&37.8&1.8&0.2&72.8&0.4&0.9&87.6&1.8&5.7&65.6&10.4\\
	CONSE \cite{norouzi_zero-shot_2013} &6.8&39.9&11.6&1.6&72.2&3.1&0.5&\tf{90.6}&1.0&0.0&\tf{91.2}&0.0\\
	CMT \cite{socher_zero-shot_2013} &8.1&21.8&11.8&7.2&49.8&12.6&0.5&90.0&1.0&1.4&85.2&2.8\\
	SSE \cite{zhang_zero-shot_2015} &2.1&36.4&4.0&8.5&46.9&14.4&8.1&82.5&14.8&0.2&78.9&0.4\\
	LATEM \cite{xian_latent_2016} &14.7&28.8&19.5&15.2&57.3&24.0&11.5&77.3&20.0&0.1&73.0&0.2\\
	ALE \cite{akata_label-embedding_2016} &21.8&33.1&26.3&23.7&62.8&34.4&14.0&81.8&23.9&4.6&73.7&8.7\\
	DEVISE \cite{frome_devise:_2013} &16.9&27.4&20.9&23.8&53.0&32.8&17.1&74.7&27.8&4.9&76.9&9.2\\
	SJE \cite{akata_evaluation_2015} &14.7&30.5&19.8&23.5&59.2&33.6&8.0&73.9&14.4&3.7&55.7&6.9\\
	ESZSL \cite{romera-paredes_embarrassingly_2015} &11.0&27.9&15.8&12.6&63.8&21.0&5.9&77.8&11.0&2.4&70.1&4.6\\
	SYNC \cite{changpinyo_synthesized_2016} &7.9&43.3&13.4&11.5&70.9&19.8&10.0&90.5&18.0&7.4&66.3&13.3\\
	SAE \cite{kodirov_semantic_2017} &8.8&18.0&11.8&7.8&54.0&13.6&1.1&82.2&2.2&0.4&80.9&0.9\\
	GFZSL \cite{verma_simple_2017} &0.0&39.6&0.0&0.0&45.7&0.0&2.5&80.1&4.8&0.0&83.3&0.0\\
	\hline
	SP-AEN \cite{chen_zero-shot_2017} & --- & --- & 24.9 & --- & --- & 34.7 & --- & --- & 23.3 & --- & --- & 13.7 \\
	PSR \cite{annadani_preserving_2018} & 20.8 & 37.2 & 26.7 & 24.6 & 54.3 & 33.9 & 20.7 & 73.8 & 32.3 & 13.5 & 51.4 & 21.4 \\   
	\hline
	AEEN  &38.9 &18.3 &24.9 &\tf{83.4} &30.6 &44.8 &\tf{95.1} &7.2 &13.4 &76.6 &10.8 &18.9\\
	AEEN (Rec)  &33.1 &40.7 &36.5 &72.5 &\tf{64.7} &\tf{68.4} &81.6 &52.5 &63.9 &51.0 &27.0 &35.4\\
	\hline
	AEEN-HO  &\tf{41.4} &18.1 &25.1 &\tf{83.4} &26.4 &40.1 &95.0 &3.5 &6.7 &\tf{77.2} &7.0 &12.8\\
	AEEN-HO (Rec)  &33.8 &\tf{46.4} & \tf{39.1} &67.9 &62.9 &65.3 &82.0 &55.6 &\tf{66.3} &55.0 &26.5 &\tf{35.7}\\
	\hline
}

\begin{figure*}[!htp]
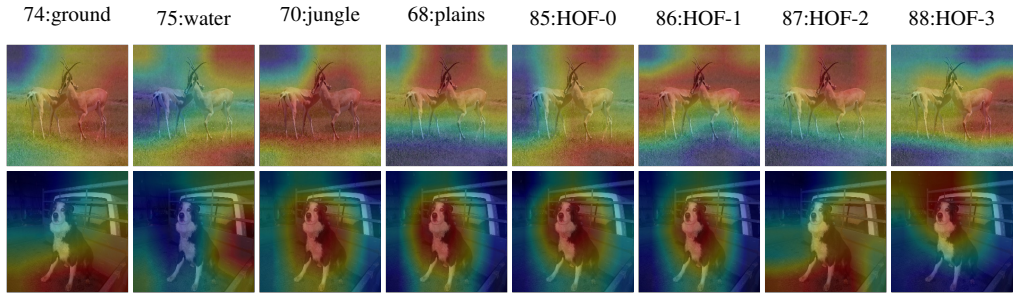

	
	\centering
	\begin{tikzpicture}[scale=0.8,transform shape] 	     
	\foreach \i/\idx/\f in {0/74/ground,1/75/water,2/70/jungle,3/68/plains,4/85/HOF-0,5/86/HOF-1,6/87/HOF-2,7/88/HOF-3}
	{
		\node[draw,inner sep=0] (0) at (\i*2.1cm,0) {\includegraphics[width=2cm]{feat-0-\idx.jpg}};
		
		\node[draw,inner sep=0] (1) at (\i*2.1cm,2.1cm) {\includegraphics[width=2cm]{feat-14-\idx.jpg}};
		\node[above=0.2cm of 1]{\idx:\f};
	}	
	\end{tikzpicture}
	\caption{First-order and high-order class attribute activation maps of the AWA2 dataset: the class attribute before and after 85 is the first-order and the high-order class attributes, respectively. We can see that the higher-order and first-order attributes complement each other. The high-order attributes also can guide the convolution map to find the discriminant region where the first-order attribute may ignore.}
	\label{fig:high-order-heat}
\end{figure*}

\begin{figure*}[!htp]
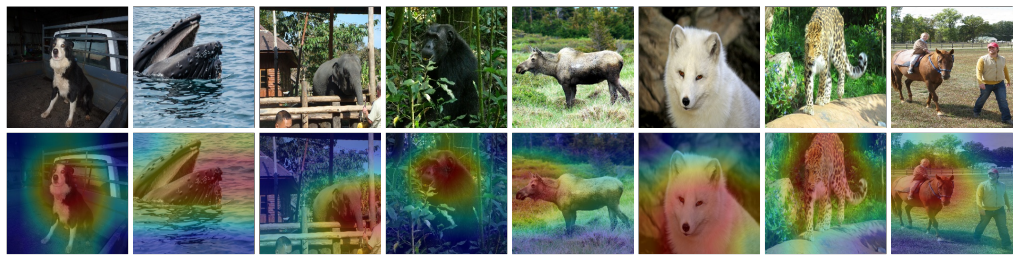

	\centering
	\begin{tikzpicture}[scale=0.8,transform shape]
	\foreach \i in {0,...,7}
	{
		
		\node[draw,inner sep=0] (1-\i) at (\i*2.1cm,2.1cm) {\includegraphics[width=2cm]{1-\i.jpg}};
		
		\node[draw,inner sep=0] (0-\i) at (\i*2.1cm,4.2cm) {\includegraphics[width=2cm]{0-\i.jpg}};
	}
	
	\end{tikzpicture}
	\caption{Average attribute activation map of AEEN-HOA on the AWA2 dataset: the images in the upper row and the lower row is the original ones and its attention maps. We can see that our approach is able to discover the irregular discriminative region of the object.}
	\label{fig:avg-heat-map}
\end{figure*}

\ssn{Comparisons with benchmarks}
To demonstrate the effectiveness of our AEEN-HOA (AEEN), we demonstrate them with 15 existing ZSL methods in Table \ref{tab:results} and \ref{tab:gzsl}, among which the results of 13 methods are the baselines reported in \cite{xian_zero-shot_2017}.

\tf{Comparisons in conventional ZSL.} In conventional ZSL setting, we follow the experiment and evaluation protocal as \cite{xian_zero-shot_2017} and reported the results on four benchmarks for both of the standard split (SS) and the proposed split (PS). The first 13 baselines from \cite{xian_zero-shot_2017} and that of next two ones are taken from \cite{chen_zero-shot_2017,annadani_preserving_2018}. We obtain our results following the identical settings for the fairness of comparisons. It can be seen that our AEEN and AEEN-HOA algorithm outperform other state-of-arts algorithms on most datasets. For example, AEEN outperforms SYNC by 16.7\% on the SS split of CUB dataset (CUB-SS), where SYNC achieves the best result in the compared methods. In addition, in the PS split of CUB dataset (CUB-PS), AEEN overpass PSR algorithm by 17.3\%. On AWA2 datasets, AEEN-HOA exceeds the best results by 6.4\% and 3.4\% for SS and PS, respectively. These results demonstrate that for images recognition in a complex background, the extraction of irregular discriminating region is very beneficial for migrating from the training classes to the test ones.

When exploring the effects of the high-order class attributes, we find that the simple off-line extracted high-order attributes help improve our algorithm further by 2\% in most cases. With an exception, we achieve about 6\% increase on the SS split of AWA2 dataset comparing AEEN-HOA with AEEN. We also observe that there is a slightly decrease of performance in CUB, which may be attributed to that the images in CUB contains a single object and simple background and there may be no such interaction between the class attributes. From the above analysis, we verify the validity of the high-order attributes for ZSL problems.

\tf{Comparisons in generalized ZSL.} When assigning the images to both of the training and test classes, our model is also comparable to other counterparts, especially with the rectifying strategy. We follow the settings of the generalized ZSL problem \cite{xian_zero-shot_2017} to report the results with the trained model on the PS split of four datasets. We can find the classification performance is biased in the training and test classes for the listed algorithms. The reason for this phenomenon is that no instances from the test classes are observed during the training process so the outputs of the training and test classes are independent of each other and not comparable during testing stage. 

With the rectifying strategy, we can well overcome the bias of the mode outputs on the training and test classes. We use a simple linear mode to select the optimal threshold parameters, and rectify the outputs on the training classes so that the training accuracy and test accuracy are as close as possible. We observe that the harmonic accuracy of our algorithms is greatly improved. In the CUB dataset, the harmonic accuracy increase from 44.8\% before rectifying to 68.4\%. As another example, the harmonic accuracy of AEEN-HOA has been greatly improved from 6.7\% before rectifying to 66.3\%. Of course, with this strategy, the harmonic accuracy of our algorithm is far more than other algorithms listed in the table.

\ssn{Attention of AEEN-HOA algorithm}
The $1\times 1$ convolutional layer generates maps with $d$ channels, where $d$ is the dimension of the class attributes.  We sample some images from the AWA2 dataset and visualize the attention map related to some attributes to obtain class attributes activation map (Figure \ref{fig:feat-heat-map}). It is surprising in the ZSL problem, our AEEN-HOA can relate the semantic objects of image to the corresponding class attributes. For example, In Figure \ref{fig:feat-heat-map}, the ground where the tiger and the dog stand, the plains where the antelopes and sheep live are marked as deeper red (attention regions). However, before training, we do not associate the position of the specific attribute of the image with the class attribute. We only use the text attribute to describe whether there is such an attribute in the image or how likely it possesses such an attribute. Our algorithm is able to accurately mark the locations of the class attributes in the image, which will aid us to understand deeply how our ZSL algorithm works.

In order to investigate how the high-order attribute activates the feature map, we show the comparison of the feature activation map of the first-order and high-order attributes on two images in Figure \ref{fig:high-order-heat}. We can see that the high-order features focus on different parts of the image, and these parts may be ignored by the first-order features. To some extent, higher-order features complement and  enhance the effects of the first-order features.

Finally, we weight the feature maps corresponding to the class attributes to obtain the average activation map of the class attributes as shown in Figure \ref{fig:avg-heat-map}, where the weights of the class attributes are softmax values \cite{zhou_learning_2015} of the class attributes matrix. We can see that AEEN-HOA can accurately find the discriminating area of the target. For example, for the rhinoceros (the 5th picture in the image), we identify whether an animal is a rhinoceros or not through its mouth rather than the body, so the color of the head of the rhinoceros appears deeper than the body in Figure \ref{fig:avg-heat-map}.  The rightmost picture in the image shows that an adult is holding a horse on which a little girl is riding. AEEN-HOA deepens the color of the first half of the horse instead of the little girl or the adult because the class of image is labeled as a horse.

\section{Conclusions}
In this paper, an adversarial erasing embedding network with the guidance of high-order attributes~(AEEN-HOA) is proposed for going further to solve the challenging ZSL/GZSL task. AEEN-HOA consists of two branches, i.e., the upper stream is capable of erasing some initially discovered regions, then the high-order attribute followed by Gaussion random projection is incorporated to represent the relationship between the class attributes. Meanwhile, the bottom stream is trained by taking the current background regions to train the same attribute. As far as we know, it is the first time of introducing erasing into the ZSL task. A class attribute activation map is proposed to visually show the relationship between class attribute features and attention map. Experiments on four standard benchmark datasets demonstrate the superiority of AEEN-HOA framework.

\bibliographystyle{abbrv}

\end{document}